\documentclass[lettersize,journal]{IEEEtran}
\usepackage{graphicx}
\usepackage{float}
\usepackage{subfig}
\usepackage[dvipsnames]{xcolor}
\usepackage{colortbl}
\usepackage{array}  
\usepackage{multicol, multirow}
\usepackage{mathtools}
\usepackage{amsfonts}
\usepackage{booktabs}

\ifCLASSOPTIONcompsoc
  \usepackage[nocompress]{cite}
\else
  \usepackage{cite}
\fi

\begin{document}

\title{RealignDiff: Boosting Text-to-Image Diffusion Model with Coarse-to-fine Semantic Re-alignment}
%
%
%
%


\author{Zutao Jiang*, Guian Fang*, Jianhua Han, Guansong Lu, \\ Hang Xu, Shengcai Liao, Xiaojun Chang, Xiaodan Liang\textsuperscript{\dag} 
\IEEEcompsocitemizethanks{
\IEEEcompsocthanksitem *~These two authors contribute equally to this work.
\IEEEcompsocthanksitem \textsuperscript{\dag}~Xiaodan Liang is the corresponding author.
\IEEEcompsocthanksitem Zutao Jiang is with Pengcheng Laboratory, Shenzhen, China. (E-mail:taozujiang@gmail.com).
\IEEEcompsocthanksitem Guian Fang and Xiaodan Liang are with Sun Yat-sen University. (E-mail:\{ fanggan@mail2.sysu.edu.cn, xdliang328@gmail.com\}).
\IEEEcompsocthanksitem Jianhua Han, Guansong Lu and Hang Xu are with Huawei Noah’s Ark Lab.
(E-mail:\{hanjianhua4@huawei.com
, luguansong@huawei.com, xu.hang@huawei.com\}).
\IEEEcompsocthanksitem  Xiaojun Chang is with the Australian Artificial Intelligence Institute, University of Technology Sydney. (E-mail:xiaojun.chang@uts.edu.au) 
\IEEEcompsocthanksitem  Shengcai Liao is with the College of Information Technology (CIT), United Arab Emirates University (UAEU), the United Arab Emirates. \\
(E-mail:scliao@ieee.org).
}
}

\markboth{Journal of \LaTeX\ Class Files,~Vol.~xx, No.~xx, xx~2024}%
{Jiang \MakeLowercase{\textit{et al.}}: RealignDiff: Boosting Text-to-Image Diffusion Model with Coarse-to-fine Semantic Re-alignment}

\IEEEtitleabstractindextext{%
\begin{abstract}
 Recent advances in text-to-image diffusion models have achieved remarkable success in generating high-quality, realistic images from textual descriptions. However, these approaches have faced challenges in precisely aligning the generated visual content with the textual concepts described in the prompts. In this paper, we propose a two-stage coarse-to-fine semantic re-alignment method, named RealignDiff, aimed at improving the alignment between text and images in text-to-image diffusion models. In the coarse semantic re-alignment phase, a novel caption reward, leveraging the BLIP-2 model, is proposed to evaluate the semantic discrepancy between the generated image caption and the given text prompt. Subsequently, the fine semantic re-alignment stage employs a local dense caption generation module and a re-weighting attention modulation module to refine the previously generated images from a local semantic view. Experimental results on the MS-COCO and ViLG-300 datasets demonstrate that the proposed two-stage coarse-to-fine semantic re-alignment method outperforms other baseline re-alignment techniques by a substantial margin in both visual quality and semantic similarity with the input prompt.
\end{abstract}

}

\maketitle

\IEEEdisplaynontitleabstractindextext

%
\IEEEpeerreviewmaketitle

\ifCLASSOPTIONcompsoc
\IEEEraisesectionheading{\section{Introduction}\label{sec:introduction}}
\else
\section{Introduction}
\label{sec:introduction}
\fi

\begin{figure*}[ht]
    \centering
    \includegraphics[width=\linewidth]{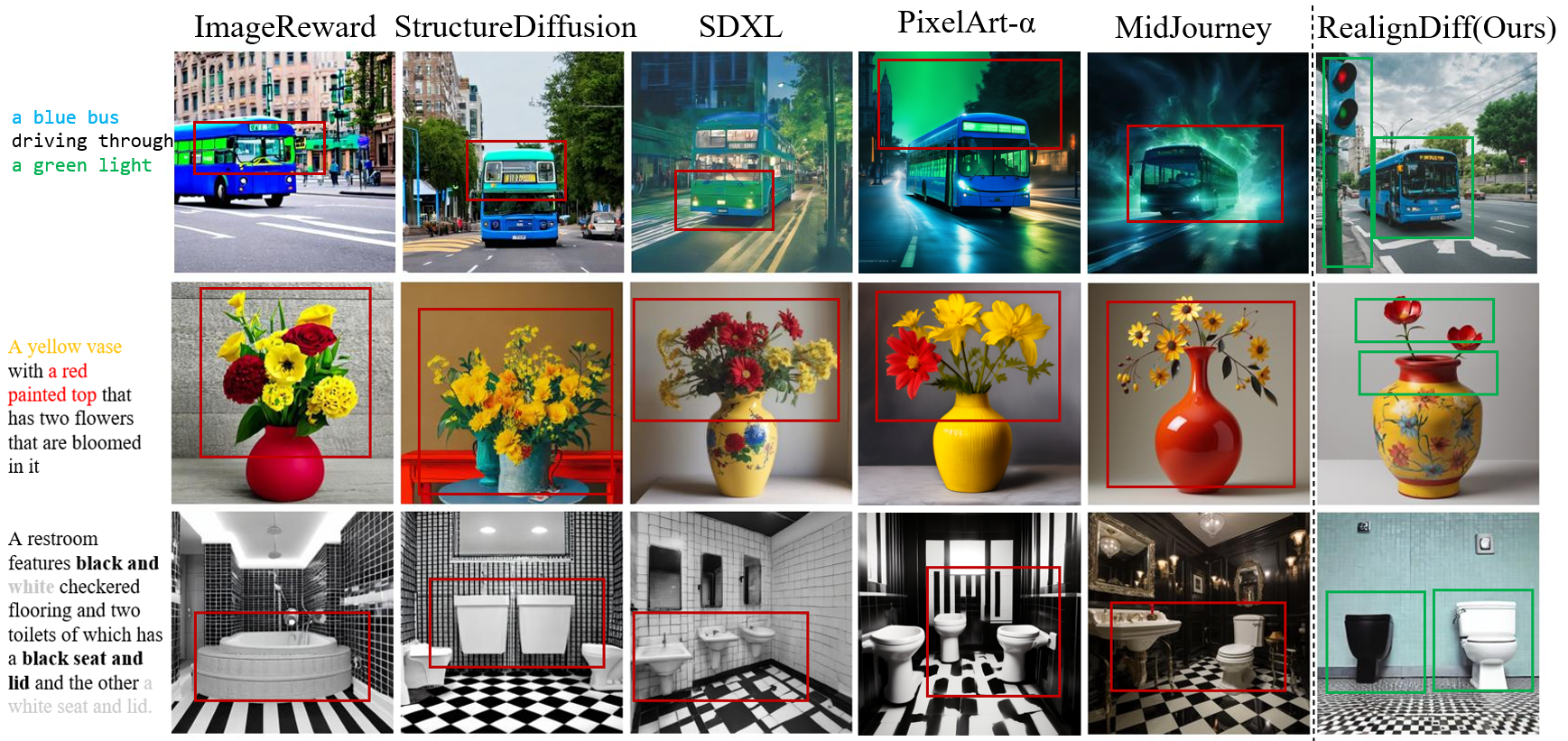}
    \caption{Visual comparison of generated images from various text-to-image diffusion models (ImageReward~\cite{xu2023imagereward}, Structure Diffusion~\cite{feng2023trainingfree}, Stable Diffusion XL~\cite{Rombach_2022_CVPR}, PixArt-$\alpha$~\cite{chen2023pixartalpha} and MidJourney\textsuperscript{\ref {midjourney}}). The motivation behind our proposed RealignDiff is to address the misalignments and semantic discrepancies observed in prior methods. From top to bottom: \textbf{Missing Main Objects} (e.g., the green traffic light in the first row is absent); \textbf{Attribute Misalignment} (e.g., the second row fails to paint red on the top of the yellow vase); \textbf{Attribute Interchange} (e.g., the third row, intended to be black and white, is not in monochrome, with the notable absence of the black toilet seat as evidence of the mix-up). RealignDiff endeavors to fix these inconsistencies, ensuring images that are more aligned with the provided textual prompts.}
    \label{fig:motivation}
\end{figure*}

\IEEEPARstart{T}{ext-to-image} diffusion models~\cite{saharia2022photorealistic, ramesh2022dalle2, Rombach_2022_CVPR, xu2022versatile} have witnessed significant advancements in recent years. These models can generate high-quality and diverse images based on the given input texts. The ability to convert textual descriptions into realistic images has enormous potential in various applications such as graphic design, computer vision, and creative writing.
Despite several text-to-image diffusion models have been deployed in real-world applications such as Imagen~\cite{saharia2022photorealistic}, DALL-E 2~\cite{ramesh2022dalle2}, Stable Diffusion~\cite{Rombach_2022_CVPR}, Midjourney\footnote{https://www.midjourney.com/\label{midjourney}},
and Versatile Diffusion~\cite{xu2022versatile}, the generated images from these models are not perfect~\cite{xu2023imagereward} as displayed in Figure \ref{fig:motivation}.
The main challenge faced by existing text-to-image diffusion models is achieving precise alignment between the generated image and the input caption. Specifically, these models often encounter difficulties in accurately capturing the attributes and relationships of the objects described in the input text. 

To tackle the issue of semantic misalignment in text-to-image diffusion models, 
some researchers have introduced pre-trained image-text models, such as CLIP~\cite{radford2021learning} and BLIP~\cite{li2022blip}, to calculate the semantic guidance~\cite{kim2022diffusionclip, li2023semantic}. Recently, ImageReward~\cite{xu2023imagereward} has also been proposed to solve both the text-image alignment and human aesthetic problems, which is trained using a comprehensive collection and annotation pipeline that leverages expert preference data.
However, as shown in Figure \ref{fig:motivation}, while ImageReward is capable of aligning images and text at a coarse level, it tends to overlook fine-grained alignment which is crucial for text-matched image generation, such as precisely aligning attributes, quantities, and relationships between objects described in the given text. To solve the fine-grained semantic alignment problem, Structure Diffusion \cite{feng2023trainingfree} incorporates language structures into the cross-attention layers. While Structure Diffusion can address the attribute binding problem, it tends to miss the main objects described in the input prompt.

In this paper, we present a novel two-stage coarse-to-fine semantic re-alignment method, called \textbf{RealignDiff}, to generate images that more accurately align with user-provided textual descriptions within text-to-image diffusion models. During the coarse semantic re-alignment stage, we propose a novel caption reward to optimize the text-to-image diffusion model from a global semantic view. Specifically, the caption reward generates a corresponding detailed caption that depicts all crucial contents in the synthetic image via a BLIP-2 model and then calculates the reward score by measuring the similarity between the generated caption and the given prompt. The elaborated caption can give more guidance about whether the surrounding concepts and context in the image are reasonable given the input text prompt. It is noteworthy that only the coarse semantic re-alignment stage may not be sufficient to capture all the desired characteristics of the generated images, especially in cases where the input texts describe complex and diverse scenes. To sense the correctness of local semantic parts, we further propose the fine semantic re-alignment. In the fine semantic re-alignment stage, we present a local dense caption generation module and a re-weighting attention modulation module from the local semantic view to refine the previously generated images. The local dense caption generation module generates the mask, detailed caption, and the corresponding likelihood score of each object appearing in the generated images. Armed with the generated detailed captions and the corresponding scores, the re-weighting attention modulation module can re-align the generated captions and the segmented parts of the generated images.

Experimental results on the MS-COCO and ViLG-300 datasets demonstrate that the proposed two-stage coarse-to-fine semantic re-alignment method outperforms other baseline re-alignment techniques by a substantial margin in both visual quality and semantic similarity with the input prompt. Our approach opens up new avenues for research in this exciting field by providing a more accurate and precise alignment mechanism that can better capture the semantic meaning of the input text and generate high-quality images. 
Our main contributions are summarized as follows:

\begin{itemize}

\item We propose a two-stage coarse-to-fine semantic re-alignment method for text-to-image diffusion models. The coarse semantic re-alignment stage ensures that the objects described in the given text appear in the generated images. The fine semantic re-alignment stage accurately captures the attributes and relationships of the objects in the input text.

\item We propose a novel caption reward and a novel local dense caption generation module. The caption reward measures the similarity between the generated caption and the given text prompt. The local dense caption generation module provides guidance regarding the attributes and spatial arrangements of objects.

\item Experimental results on the MS-COCO~\cite{lin2014microsoft} and ViLG-300~\cite{feng2023ernie} datasets demonstrate that \textbf{RealignDiff} can better align the semantics of the generated image from the text-to-image diffusion model with the given text prompt, achieving the best performance compared to other baseline methods.
\end{itemize}
\section{Related Work}
\label{sec:related_work}

\subsection{Text-to-Image Generation.} Text-to-image generation aims to generate images given input text descriptions. Along with the progress on generative models, including generative adversarial networks (GANs ~\cite{goodfellow2014gan}), auto-regressive model ~\cite{vaswani2017attention} and diffusion model~\cite{ho2020ddpm}, there are numbers of works for text-to-image generation. Among them, GANs are first adopted for text-to-image generation~\cite{GAN-INT-CLS} and later many GAN-based models are proposed for better visual fidelity and caption similarity~\cite{zhang2018stackgan++,xu2018attngan,li2019controllable,sisgan,zhu2019dmgan,tao2020dfgan,ye2021xmcgan,kang2023scaling,sauer2023stylegan}. 
However, GANs suffer from the well-known problem of mode-collapse and unstable training processes. 
To solve these problems, another line of works explore applying Transformer-based auto-regressive model for text-to-image generation~\cite{dalle,ding2021cogview,esser2021imagebart,ding2022cogview2,zhang2021m6-ufc,lee2022rq-vae,chang2023muse} with a discrete VAE~\cite{2017Neural,vqvae2,vqgan} model for tokenizing the input images and a Transformer~\cite{vaswani2017attention} model for fitting the joint distribution of text tokens and image tokens. 
Recent works adopt diffusion model for text-to-image generation~\cite{nichol2021glide,ho2022cascaded-diffusion,ramesh2022dalle2,saharia2022photorealistic,Rombach_2022_CVPR,xu2022versatile},
which learns to predict the added noise of noised images and generates images from pure noise by iteratively predict added noise and remove it. 
Among them, in order to reduce the computational overhead of large-scale text-to-image generation models, Stable Diffusion ~\cite{Rombach_2022_CVPR} proposed to first encode the input images as low-dimension latent codes and then adopt a diffusion model to generate these latent codes conditioned on the input texts. Although significant progress in high-quality text-to-image generation has been achieved, problems including misalignment with human preference and misalignment with input texts still remain to be solved.


\subsection{Alignment of Text-to-Image Generation Models.}
Some works~\cite{hao2022optimizing,lee2023aligning,wu2023better,xu2023imagereward,dong2023raft} are proposed to align a text-to-image generation model with human preference and aesthetic quality. 
\cite{lee2023aligning} first learn a reward model with the human
feedback assessing model outputs and then finetune a text-to-image model by maximizing reward-weighted likelihood to improve image-text alignment. 
Similarly, \cite{wu2023better} took the human aesthetic preference into account and proposed to learn a human preference reward model. ImageReward~\cite{xu2023imagereward} proposed a general-purpose text-to-image human preference reward model, covering text-image alignment, body problems, human aesthetics, toxicity, and biases. 
Promptist~\cite{hao2022optimizing} proposed prompt adaptation, \textit{i.e.}, training a language model to generate a better prompt given the origin prompt. They utilize the CLIP model and aesthetic predictor model as the reward model and perform supervised fine-tuning under the reinforcement
learning paradigm.
On the other hand, to circumvent the problems of  inefficiencies and instabilities of Reinforcement
Learning from Human Feedback (RLHF~\cite{ouyang2022training}), \cite{dong2023raft} introduce Reward ranked Fine-Tuning (RAFT) to align generative models more effectively.
However, RAFT is prone to overfitting as the number of iterations increases. More recently, Xu et al.~\cite{xu2023imagereward} have developed reward feedback learning (ReFL) to optimize text-to-image diffusion models against a reward function, which has demonstrated its effectiveness in achieving better alignment.
However, the existing reward models do not take both coarse-grained and fine-grained image-text semantic alignment into account. In this paper, we propose RealignDiff to improve the alignment between text and images in text-to-image diffusion models from the global and local views.

\section{Method}
\label{sec:method}

\begin{figure*}[t!]
\centering
\includegraphics[width=1.0\linewidth]{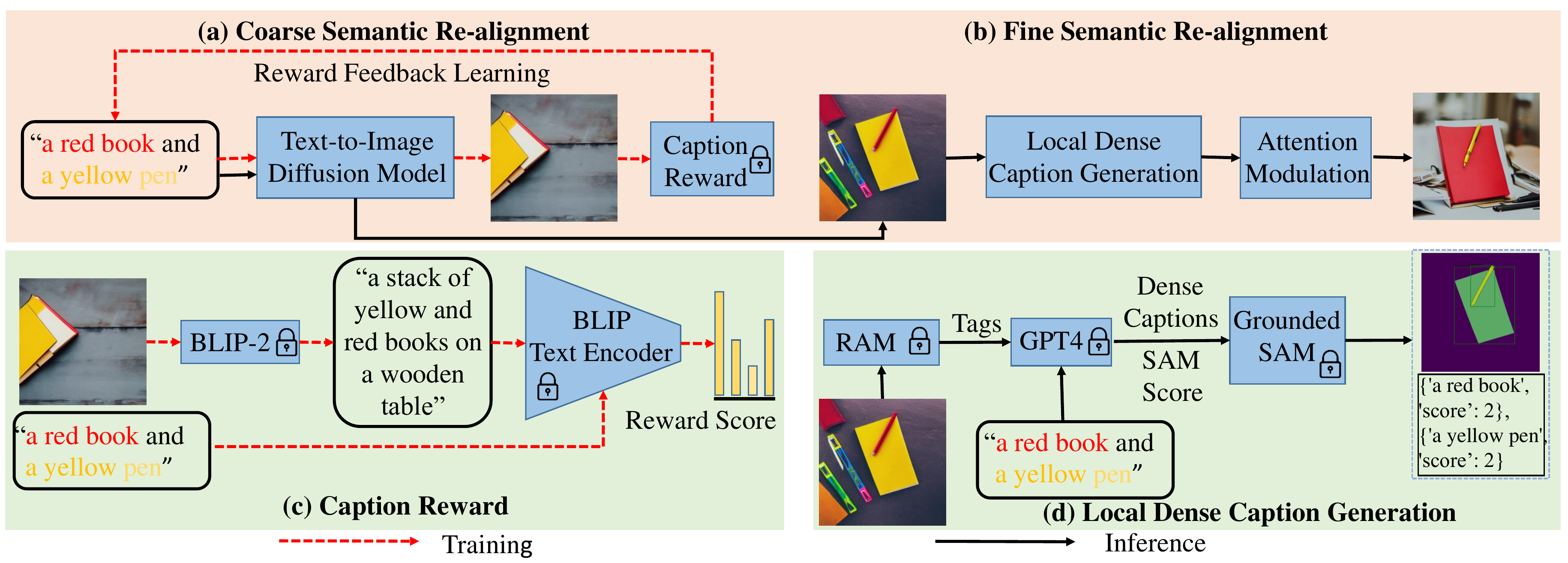}
\caption{The framework of our RealignDiff approach.
(a) Coarse Semantic Re-alignment enables the objects described in the given text to appear in the generated images. (b) Fine Semantic Re-alignment accurately captures the attributes and relationships of the objects. 
(c) Caption Reward measures the similarity between the generated caption and the given prompt. (d) The local dense caption generation module provides guidance regarding the attributes and spatial arrangements of objects within the fine semantic re-alignment stage.
}
\label{fig:framework}
\end{figure*}

Figure \ref{fig:framework} shows the pipeline of our RealignDiff approach for boosting the text-to-image diffusion models.
In this section, we first introduce the preliminary knowledge of the text-to-image diffusion model. Then we present the coarse semantic re-alignment method, including the caption reward and reward feedback learning framework. Finally, we introduce the fine semantic re-alignment method, including the local dense caption generation and the re-weighting attention modulation modules. 

\subsection{Preliminary: Text-to-image diffusion model} 

Given an image sampled from the real image distribution $x_0 \sim q(x_0)$, diffusion models first produce a Markov chain of latent variables $x_1,...,x_T$ by progressively adding Gaussian noise to the image according to some variance schedule given by $\beta_t$ as follows:
\begin{equation}
    q\left(x_t \mid x_{t-1}\right) = \mathcal{N}\left(\sqrt{1-\beta_t} x_{t-1}, \beta_t I\right),
\end{equation}
and then learn a model 
to approximate the true posterior:
\begin{equation}
        p_\theta\left(x_{t-1} \mid x_t\right) = \mathcal{N}\left(\mu_\theta\left(x_t, t\right), \Sigma_\theta\left(x_t, t\right)\right),
\end{equation}
to perform the reverse denoising process for image generation: starting from a random noise $x_T \sim \mathcal{N}(0, I)$ and gradually reducing the noise to finally get a real image $x_0$. While a tractable variational lower-bound $\mathcal{L}_{VLB}$ on $\log p_\theta(x_0)$ can be used to optimize $\mu_\theta$ and $\Sigma_\theta$, to achieve better results, \cite{ho2020ddpm} instead adopt a denoising network $\epsilon_\theta(x_t, t)$ which predicts the added noise of a noisy image $x_t \sim q(x_t | x_0)$ and adopts the following training objective:
\begin{equation}
    \mathcal{L} = \mathbb{E}_{x_0 \sim q\left(x_0\right), \epsilon \sim \mathcal{N}(0, I), t \sim[1,T]}\left\|\epsilon-\epsilon_\theta\left(x_t, t\right)\right\|^2,
\end{equation}
where $t$ is uniformly sampled from $\{1,...,T\}$. For a text-to-image generation, the denoising network receives the input text $t_p$ as extra conditional input and is denoted as $\epsilon_\theta(x_t, t_p, t)$.

We adopt Stable Diffusion~\cite{Rombach_2022_CVPR} as our baseline text-to-image generation model. In this model, a real image is first down-sampled 8 times as a lower-dimension latent code $x_0$ with an autoencoder model and the denoising network $\epsilon_\theta(x_t, t_p, t)$ is parameterized as a Unet~\cite{ronneberger2015u} network, where embedding of time step $t$ is injected with adaptive normalization layers and embedding of input text $t_p$ is injected with cross-attention layers. However, the Stable Diffusion model fails to perform precise alignment between the text concept and generated images since it's trained only with the global alignment between the text and images.

\subsection{Coarse-to-fine Semantic Re-alignment~}

In this subsection, we first introduce the coarse semantic re-alignment method and then present the fine semantic re-alignment method.

\subsubsection{Coarse Semantic Re-alignment.} 

To ensure the objects described in the given text appear in the generated image, we propose the coarse semantic re-alignment method, including the caption reward and the reward feedback learning framework.

\noindent \textbf{Caption Reward.}
The caption reward is proposed to improve the consistency between the synthetic caption of the generated image and the given text prompt. 
Specifically, given an image generated by a text-to-image diffusion model, we first obtain the corresponding caption $t_g$ using the pre-trained Blip-2 model. Then we compute the similarity between the embeddings of generated caption $t_g$ and the corresponding text prompt $t_p$ as our caption reward score. Note that we utilize a pre-trained BLIP-2 \cite{li2023blip} text encoder  ${{\mathop{\rm f}\nolimits} _{enc}}( \cdot )$ to convert the captions into the text embeddings. 
Formally, the caption reward score ${\bf{R}}_{cap}$ can be calculated as follows:
\begin{equation}
{{\bf{R}}_{cap}} = \frac{{{{\mathop{\rm f}\nolimits} _{enc}}({t_g}) \cdot {{\mathop{\rm f}\nolimits} _{enc}}({t_p})}}{{\left\| {{{\mathop{\rm f}\nolimits} _{enc}}({t_g})} \right\|\left\| {{{\mathop{\rm f}\nolimits} _{enc}}({t_p})} \right\|}}.
\label{eq:caption_reward}
\end{equation}

Note that while the caption reward can effectively promote consistency between the generated captions and input prompts, it may not capture all the desired characteristics of the generated images, especially in cases where the input texts describe complex and diverse scenes.

\noindent \textbf{Reward Feedback Learning.}
Reward Feedback Learning (ReFL) is designed to optimize text-to-image diffusion models by leveraging a reward function. Within this framework, a caption reward is incorporated to enable coarse semantic re-alignment. ReFL facilitates the direct optimization of text-to-image diffusion models by back-propagating gradients to a randomly selected intermediate step $t$ during the denoising process. The rationale behind this random selection of $t$ is significant: solely retaining the gradient information from the last denoising step leads to pronounced training instability and suboptimal results. Instead of progressively reducing noise to generate an image ${x_0}$ from an intermediate state ${x_t}$ via a sequential transformation process ${x_t} \to {x_{t - 1}} \to \ldots \to {x_0}$, ReFL employs an alternative approach. It directly predicts ${x_0}^\prime$ from ${x_t}$ using the transformation ${x_t} \to {x_0}^\prime$ during the fine-tuning of text-to-image diffusion models. This method is grounded in the insightful observation that the caption reward scores for generations ${x_0}^\prime$ after a sufficient number of denoising steps (typically, $t \ge 30$), provide effective feedback for improving model performance.

To address the challenges of rapid overfitting and to enhance stability during fine-tuning, a re-weighting scheme is applied to the ReFL loss, along with regularization using the pre-training loss. The overall loss function is defined as:

\begin{equation}
\begin{array}{l}
\mathcal{L}_{\text{total}} = \lambda \mathcal{L}_{\text{reward}} + \mathcal{L}_{\text{pre}} \\
 =\lambda \phi ({\bf{R}_{cap}}({t_p},{g_\theta }({t_p}))) +  \\
\mathbb{E}_{x_0 \sim q\left(x_0\right), \epsilon \sim \mathcal{N}(0, I), t \sim[1,T]}\left\|\epsilon-\epsilon_\theta\left(x_t, t_p, t\right)\right\|^2,
\end{array}
\end{equation}
where $\lambda$ is a weighting factor, $\phi$ is the ReLU operation, $\bf{R}_{cap}$ denotes the caption reward score, $\theta$ represents the parameters of the text-to-image diffusion models, while ${g_\theta }({t_p})$ denotes the generated image produced by the text-to-image diffusion models with parameters $\theta$, corresponding to the text prompt $t_p$. This formulation underscores the essential role of the ReFL loss in optimizing the model's performance with respect to semantic alignment.

\begin{figure*}
    \centering
    \includegraphics[width=1.0\textwidth]{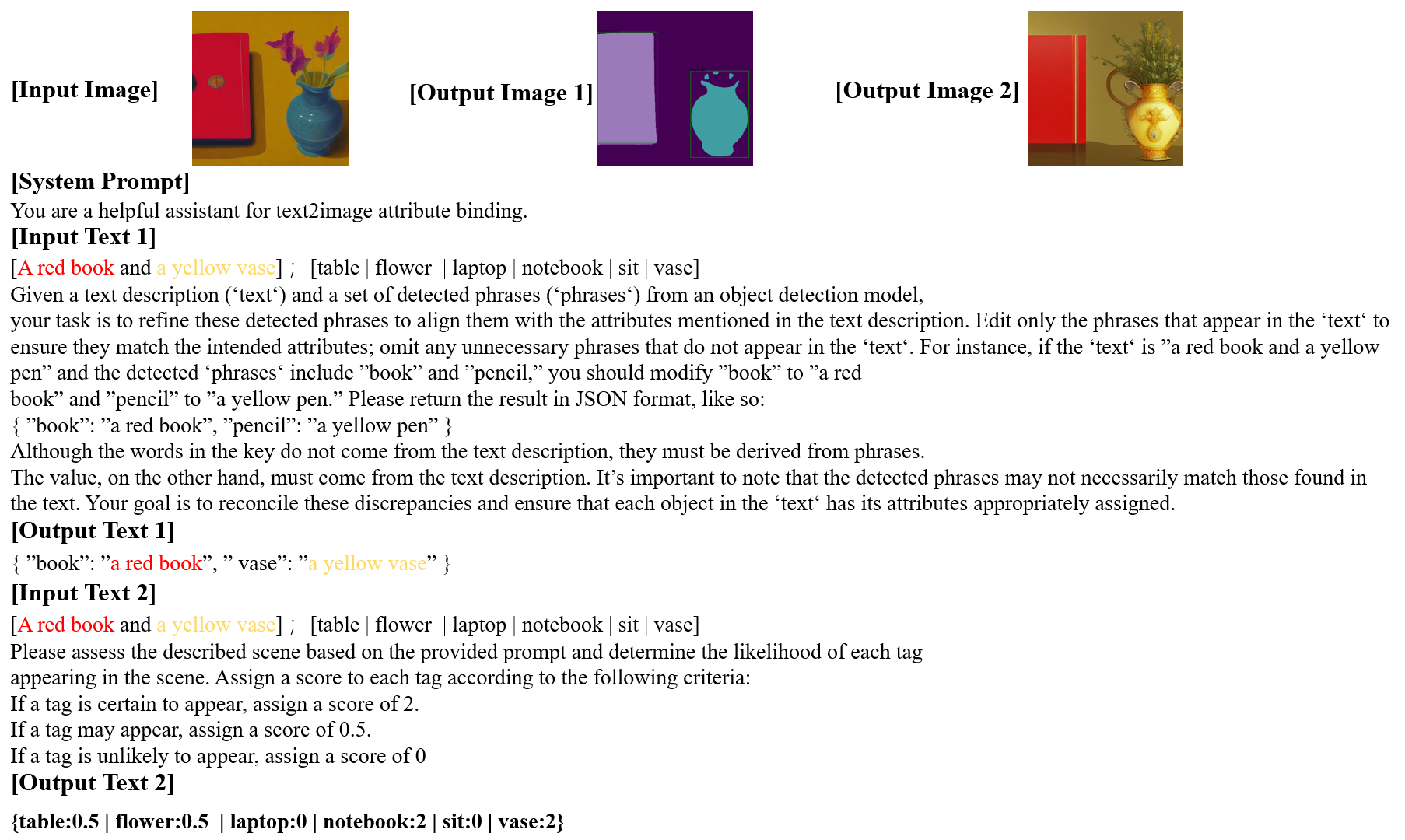}
    \caption{Input Image is aligned at a coarse level, focusing on objects. Output Image 1 illustrates the RAM process, also generating phases for text input. Text outputs 1 and 2 provide essential parameters (aligned attributes, weighted granularity) for the final generation, leading to output image 2 through fine-grained alignment.}
    \label{fig:proc}
\end{figure*}

\subsubsection{Fine Semantic Re-alignment} 
In this subsection, we present a training-free method for achieving fine-grained semantic re-alignment. Our objective is to accurately capture the attributes and relationships of the objects described in the input text. This method encompasses two key components: the local dense caption and re-weighting attention modulation.

\noindent \textbf{Local Dense Caption Generation.}
The local dense caption generation module is designed to concentrate on the specific details within the generated images and assess their alignment with the provided text descriptions from a local perspective. This method fundamentally tackles two crucial objectives: 1) Ascertaining whether the objects depicted in the generated image are consistent with the textual descriptions. 2) providing comprehensive and precise captions for the objects depicted within the generated image.

Specifically, the local dense caption generation module first 
recognizes the objects in the previously generated images using an off-the-shelf image tagging model, \textit{i.e.}, Recognize Anything Model (RAM) \cite{zhang2023recognize}. Subsequently, given the provided prompt and the image tags, the large language model, \textit{i.e.}, GPT-4 \cite{openai2023gpt4}, is utilized to assess the likelihood score $\{ {s_i}\} _{i = 1}^n$ and provide the local detailed descriptions $\{ {l_i}\} _{i = 1}^n$ for each recognized object. The score of each object is assigned based on the likelihood 
of each category label of the object appearing in the scene, which can be summarized as follows: 

\begin{equation}
  s_i=\left\{
\begin{array}{rcl}
2\text{,}       &      & \text{$c$ is certain to appear in the scene.}\\
0.5\text{,}     &      & \text{$c$  may appear in the scene.}\\
0\text{,}    &      & \text{$c$  is unlikely to appear in the scene.}\\
\end{array} \right.  
\end{equation}
where $c$ denotes the category label of the object.

Consider the text prompt `a red book and a yellow pen' as an example. We first utilize a fine-tuned text-to-image diffusion model to generate the image. Subsequently, RAM is employed to identify the objects within this image. If the object tag is `book' or `pen', GPT-4 determines that these objects are certain to appear in the scene, assigning a score of 2. Conversely, if the object tag is `sign' or `banana', GPT-4 deems these objects unlikely to appear, thus assigning a score of 0. For an object tag like `desk', which GPT-4 considers as possibly appearing in the scene, a score of 0.5 is assigned. The example of using GPT-4 is shown in Figure \ref{fig:proc}. Our extensive experiments have demonstrated GPT-4's proficiency in accurately performing such tasks.

By assigning scores in this way, we can obtain a likelihood score $\{ {s_i}\} _{i = 1}^n$ for each recognized object. 
After obtaining the local dense caption and the corresponding likelihood score, we use the off-the-shelf segmentation model, \textit{i.e.}, Grounded Semantic Segmentation anything (Grounded-SAM)\cite{liu2023grounding} to obtain the object masks $\{ {m_i}\} _{i = 1}^n$. The local dense caption and the object mask can provide guidance regarding the attributes and layout of objects with the re-weighting attention modulation method.


\noindent \textbf{Re-weighting Attention Modulation.} 
The re-weighting attention strategy plays a crucial role in controlling how specific tokens influence the resulting image. By adjusting the influence based on reward scores, more relevant semantic cues can dominate the image generation process, improving the semantic alignment between the text and the generated image. Building on this concept, we propose a re-weighting attention modulation module. Given a set of detailed local caption $\{ {l_i}\} _{i = 1}^n$ and corresponding object masks $\{ {m_i}\} _{i = 1}^n$, the module ensures that objects appear in the correct regions based on their likelihood scores $\{ {s_i}\} _{i = 1}^n$. 

Specifically, the original attention maps $A\in\mathbb{R}^{{\lvert queries \lvert} \times {\lvert keys \lvert}}$ is defined as below:
\begin{equation}
A = \text{softmax}\left(\frac{QK^T}{\sqrt{d}} \right),  
\end{equation}
where $Q$ represents the query value, which is mapped from image features, while $K$ denotes the key value, derived from text features. The term $d$ is the length of the key and query features. Based on the attention map $A$, the image features can be updated, referencing the text features.

Our re-weighting attention modulation method modulates the original attention maps $A$ as follows:
\begin{equation}
\begin{array}{l}
A' = \text{softmax}\left(\frac{QK^T + S \odot M}{\sqrt{d}} \right), \\
M = \lambda_t \cdot  R \odot M_{\text{pos}} \odot (1 - B) \\
    - \lambda_t \cdot (1 - R)  \odot M_{\text{neg}} \odot (1 - B),
\end{array}
\end{equation}
where $\odot$ denotes the Hadamard product. $\lambda_t$ is a scalar, proportional to the timestep $t$, to adjust the degree of modulation. 
$S$ represents the re-weighting score matrix, which can be obtained through the likelihood score $\{ {s_i}\} _{i = 1}^n$. The score matrix $S_i$ for each $i$ is constructed based on the elements of object masks $\{ {m_i}\} _{i = 1}^n$ 
and corresponding likelihood scores $\{ {s_i}\} _{i = 1}^n$. The rule for constructing $S_i$ is as follows:
\begin{equation}
S_i = \left[ S_{ijk} \right] \quad \text{where} \quad S_{ijk} = 
\begin{cases} 
s_i & \text{if } m_{ijk} > 0, \\
1 & \text{if } m_{ijk} \leq 0,
\end{cases} 
\end{equation}
where $j$ and $k$ represent the row and column indices in $S_i$ and $m_i $ respectively.
This formulation applies to each matrix $S_i$ in the set $ \{ S_i \} _{i = 1}^n$. We use \textit{Grounded-SAM} to obtain the object masks $\{ {m_i}\} _{i = 1}^n$. 
$R$ is a boolean mask vector where each element corresponds to a token in the text features. $R_i > 0$ indicates that the text token is activated at position $i$.
$B = QK^T$ denotes the similarity score between the query and key. 
$M_{\text{pos}}$ and $M_{\text{neg}}$ can be calculated as:
\begin{equation}
\begin{array}{l}
M_{\text{pos}} = \max(QK^T) - QK^T, \\
M_{\text{neg}} = QK^T - \min(QK^T), 
\end{array}
\end{equation}
where $M_{\text{pos}}$ denotes the maximum and $M_{\text{neg}}$ represents minimum values. With our re-weighting attention modulation, the model is guided to focus more attention on the important tokens, corresponding to the generated local dense captions. Therefore, our method can refine the previously generated images for better semantic alignment with the text prompts.

\section{Experiments}
\label{sec:experiments}

\begin{table*}[ht]
\footnotesize
\centering
\caption{Quantitative comparison of different methods on the MS-COCO~\cite{lin2014microsoft}, ABC-6K~\cite{feng2023trainingfree}, CC-500~\cite{feng2023trainingfree}, and ViLG-300~\cite{feng2023ernievilg} datasets.}
\begin{tabular}{c|c|cccccc}
\hline

    \multirow{2}{*}{Dataset} & \multirow{2}{*}{Method} & \multirow{2}{*}{FID$\downarrow$} & \multirow{2}{*}{CLIP$\uparrow$} & \multirow{2}{*}{TIFA $\uparrow$} & \multicolumn{2}{c}{Human Study} \\ \cline{6-7}
    & & & & & Alignment $\uparrow$ & Fidelity $\uparrow$ \\

     \hline

\multirow{8}{*}{MS-COCO~\cite{lin2014microsoft}} 
               & SD-v1.5~\cite{Rombach_2022_CVPR}& 13.7599 & 0.1626 & 0.78 & 8.9\% & 2.1\% \\
               & SD-XL~\cite{Rombach_2022_CVPR} & 7.0864 &  0.3578 & 0.84 & 14.9\%&  19.8\%\\
               & DeepFloyd-IF~\cite{shonenkov2023deepfloyd} & 7.5431 &  0.3433 & 0.87 &  16.1\%&  16.7\% \\
               & Imagereward~\cite{xu2023imagereward} &12.7248 &0.1587 & 0.74 & 7.5\%&  5.8\%\\        
               & DenseDiffusion~\cite{kim2023dense} &8.3359 &0.1585 & 0.80 & 8.7\%&  11.9\%\\
               & Promptist~\cite{hao2022optimizing} &8.0351 &0.1627 & 0.79 & 11.2\%&  9.1\%\\
               & StructureDiffusion~\cite{feng2023trainingfree} &8.7603 &0.3279 & 0.85  & 13.6\%&  14.3\%\\
               & PixArt-$\alpha$~\cite{chen2023pixartalpha} &  7.9378&  0.3449 & 0.84 & $-$ &  $-$ \\
               & RealignDiff (Ours) &\textbf{6.9617} &\textbf{0.3767} & \textbf{0.89} & \textbf{19.1\%} & \textbf{20.3\%}\\
\hline
\multirow{8}{*}{ABC-6K~\cite{feng2023trainingfree}} 
               & SD-v1.5~\cite{Rombach_2022_CVPR}& 13.4539 & 0.1620 & 0.75 & 9.2\% & 8.7\% \\
               & SD-XL~\cite{Rombach_2022_CVPR} &6.7145  &  0.3531 & 0.84 & 14.4\% &  18.4\% \\
               & DeepFloyd-IF~\cite{shonenkov2023deepfloyd} & 7.3319 &  0.3399 & 0.86 &  15.7\% &  15.0\% \\
               & Imagereward~\cite{xu2023imagereward} &12.4287 &0.1592 & 0.71 & 9.8\% &  6.8\% \\       
               & DenseDiffusion~\cite{kim2023dense} &8.1301 &0.1604 & 0.79 & 7.1\% &  11.1\% \\
               & Promptist~\cite{hao2022optimizing} &8.0352 &0.1636 & 0.77 & 11.6\% &  9.3\% \\
               & StructureDiffusion~\cite{feng2023trainingfree} &8.6598 &0.3301 & 0.83 & 12.8\% &  12.2\% \\
               & PixArt-$\alpha$~\cite{chen2023pixartalpha} &  7.7364&  0.3391 & 0.83 & $-$ &  $-$\\
               & RealignDiff (Ours) &\textbf{6.5623} &\textbf{0.3782} & \textbf{0.88} & \textbf{19.4\%} & \textbf{18.5\%} \\

\hline
   \multirow{8}{*}{CC-500~\cite{feng2023trainingfree}} 
               & SD-v1.5~\cite{Rombach_2022_CVPR}& 13.9598 & 0.1615 & 0.77 & 6.8\% & 4.0\% \\
               & SD-XL~\cite{Rombach_2022_CVPR} &7.2364  &  0.3498 & 0.83 & 16.5\% & 15.4\% \\
               & DeepFloyd-IF~\cite{shonenkov2023deepfloyd} & 7.5494 &  0.3614 & 0.85 & 20.1\% & 9.6\% \\
               & Imagereward~\cite{xu2023imagereward} &12.8246 &0.1582 & 0.74 & 5.3\% & 1.9\% \\      
               & DenseDiffusion~\cite{kim2023dense} &8.1353 &0.1635 & 0.82 & 10.1\% & 18.0\% \\
               & Promptist~\cite{hao2022optimizing} &8.4352 &0.1593 & 0.77 & 7.3\% & 14.4\% \\
               & StructureDiffusion~\cite{feng2023trainingfree} &8.8604 &0.3281 & 0.85 & 13.1\% & 12.3\% \\
                 &  PixArt-$\alpha$~\cite{chen2023pixartalpha} &  8.5431&  0.3211 & 0.82 & $-$ &  $-$\\
               & RealignDiff (Ours) &\textbf{7.1641} &\textbf{0.3761} & \textbf{0.90} & \textbf{20.8\%} & \textbf{24.4\%} \\

\hline

        \multirow{8}{*}{ViLG-300~\cite{feng2023ernievilg}} 
               &  SD-v1.5~\cite{Rombach_2022_CVPR}&15.4943 & 0.1957 & 0.76  &  10.1\%&  2.4\%\\
      &    SD-XL~\cite{Rombach_2022_CVPR} &  7.9753&  0.4213 & 0.83 & 14.9\%&  20.0\%\\
   &  DeepFloyd-IF~\cite{shonenkov2023deepfloyd} & 8.1541 &  0.4349 & 0.86 &  16.1\%&  16.7\% \\
      &       Imagereward~\cite{xu2023imagereward} & 13.6488& 0.1906 &  0.75 &  7.5\%&  4.8\%\\     
       &       DenseDiffusion~\cite{kim2023dense} & 8.6412&  0.1907 &  0.81&  8.7\%&  11.9\%\\
        &     Promptist~\cite{hao2022optimizing} &9.2419  & 0.1962 &  0.77 &  11.2\%&  9.5\%\\
   &StructureDiffusion~\cite{feng2023trainingfree} &9.1514 &  0.3936 &  0.86 & 13.6\%&  14.3\%\\
             &    PixArt-$\alpha$~\cite{chen2023pixartalpha} &  8.4496&  0.4015 & 0.84 & $-$&  $-$\\
           &      RealignDiff (Ours) &  \textbf{7.2311}&\textbf{0.4527} & \textbf{0.90} & \textbf{17.9\%} & \textbf{20.4\%}    \\
    \hline
\end{tabular}
\label{tab:refiner}
\end{table*}

\subsection{Experimental Setup}

\textbf{Datasets.} 
Our approach is trained on the MS-COCO~\cite{lin2014microsoft} and ViLG-300~\cite{feng2023ernie} datasets. The MS-COCO dataset comprises 82,783 training and 40,504 validation text-image pairs. We split the ViLG300 dataset into 80\% for the training set and 20\% for the test set. It is noteworthy that only the image captions from the training subset of the MS-COCO dataset and the ViLG-300 dataset are utilized for fine-tuning the model. For the evaluation, we have randomly selected 5,000 image captions from the validation subset of the MS-COCO dataset. Furthermore, our approach is also evaluated on the ABC-6K~\cite{feng2023trainingfree} and CC-500~\cite{feng2023trainingfree}. The ABC-6K, derived from natural prompts within MS-COCO, each contains a minimum of two color descriptors modifying distinct objects. In contrast, the CC-500 consists of natural compositional prompts but primarily features simpler prompts that combine two concepts. These prompts follow sentence structures like “a red car and a pink elephant”, pairing objects with their respective attribute descriptors.


\noindent \textbf{Evaluation Metrics.} We adopt two metrics to measure the semantic consistency between the generated images and the input text prompts: CLIP~\cite{radford2021learning,hessel2022clipscore}and TIFA~\cite{hu2023tifa} scores. 
The higher the CLIP and TIFA scores, the better the semantic consistency. TIFA score uses a VQA method to evaluate alignment.
Furthermore, the quality of generated images was assessed using the Frechet Inception Distance (FID)\footnote{https://github.com/mseitzer/pytorch-fid} \cite{heusel2018gans}, where a lower FID score indicates better image quality.  We also conducted a human study to gauge the semantic alignment of the generated images with their corresponding textual prompts. Participants in the study were presented with sets of images synthesized by the different text-to-image diffusion models alongside the input prompts that guided their generation. 
They were instructed to choose the order of different results in terms of the \textbf{alignment} and \textbf{fidelity} metrics. The alignment score measures the semantic consistency between the generated images and the input prompts. The fidelity score measures the quality of the generated images. We use the average rank from different participants as the final scores. 
This study collected a total of $100$ human evaluation results. 


\noindent \textbf{Implementation Details.} 
Our algorithm is implemented in PyTorch. All experiments are conducted on servers equipped with eight Nvidia A100 GPUs, each with 40 GB of memory, and an AMD EPYC 7742 CPU running at 2.30 GHz. We adopt the Stable Diffusion v1.5~\cite{Rombach_2022_CVPR} as the foundational generative model and proceed to fine-tune it. We set a learning rate at 1e-5 and utilize a cumulative batch size of 128. All training and evaluations are conducted at a resolution of 512x512. We chose our fine-tuned checkpoint based on early stopping criteria to avoid overfitting and ‘reward hacking’, leading to performance degradation. The training was stopped after approximately 947 iterations when further improvements on validation metrics ceased. For each generation task, we set the random seed to 42 and generate images with a resolution of 512x512 pixels. The model is fine-tuned using half-precision floating-point numbers. For the ReFL algorithm, we configure the settings with $\lambda=1 e-3$, and $T=50$.


\subsection{Comparison Against Baselines}
In this section, we conduct a comparative assessment of the proposed RealignDiff model against eight state-of-the-art text-to-image diffusion models. These include SD-v1.5~\cite{Rombach_2022_CVPR}, SD-XL~\cite{Rombach_2022_CVPR}, DeepFloyd-IF~\cite{shonenkov2023deepfloyd}, PixArt-$\alpha$~\cite{chen2023pixartalpha},  DenseDiffusion~\cite{kim2023dense}, Imagereward~\cite{xu2023imagereward}, Promptist~\cite{hao2022optimizing}, and StructureDiffusion~\cite{feng2023trainingfree}. Table \ref{tab:refiner} displays the quantitative comparison results of different methods on the MS-COCO, ABC-6K, CC-500 and ViLG-300 datasets.

\begin{figure*}
\centering
\includegraphics[width=1\linewidth]{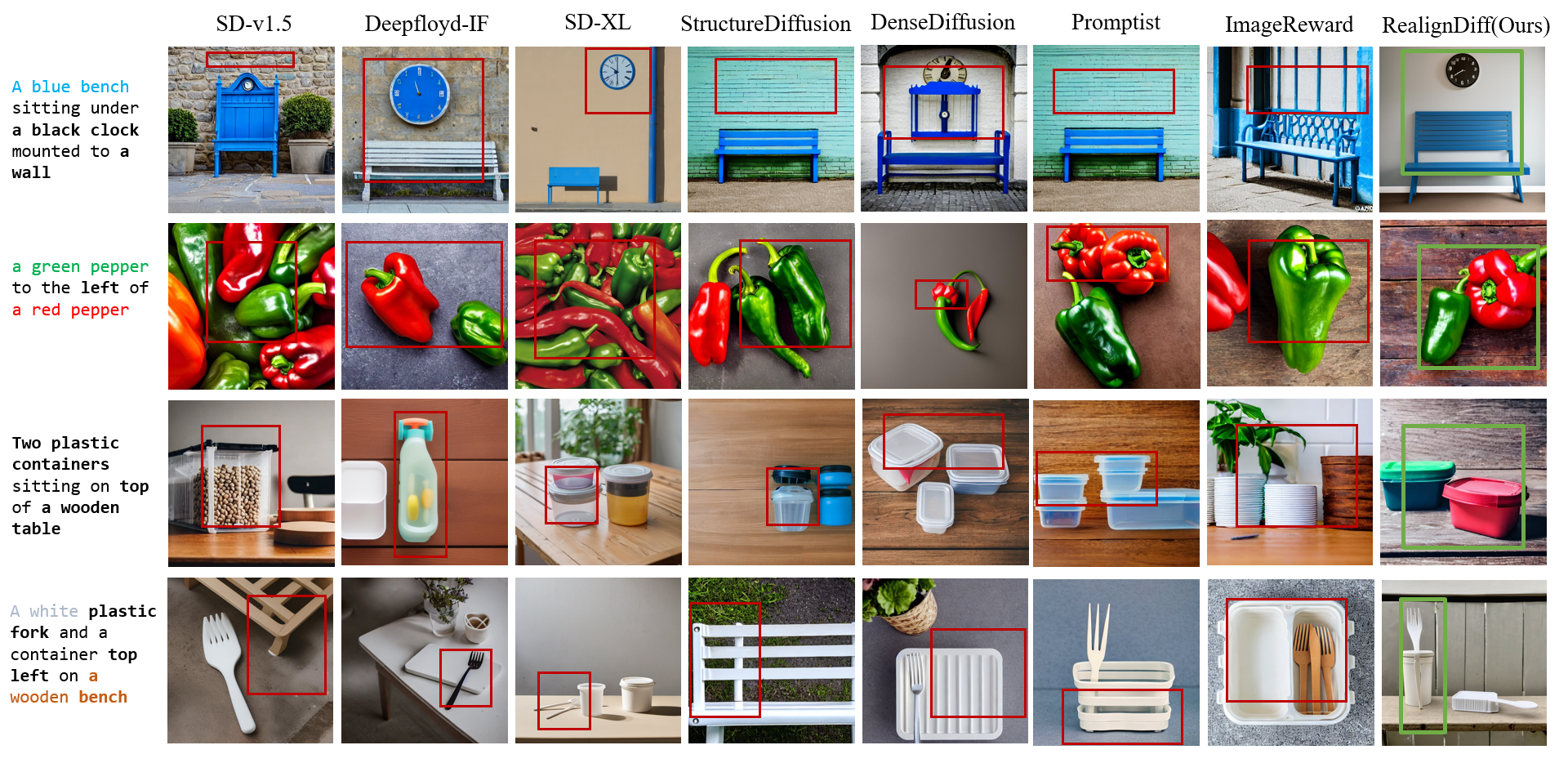}
\caption{Qualitative comparison of different methods. Our method achieves the best performance regarding the quantity of objects, leakage of attributes, and the binding of attributes. More cases are provided in the Appendix.}
\label{fig:result}
\end{figure*}

\begin{figure*}[]
	\centering  
 
	\subfloat[A cat laying \textcolor{red}{next to} a stainless \textcolor{red}{steel bowl}.]{
		\includegraphics[width=0.95\linewidth]{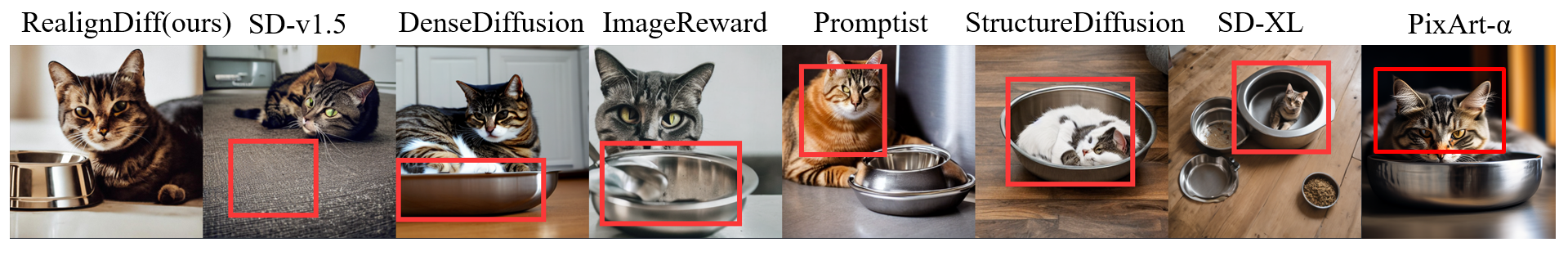} } \\
  \vspace{-2mm}
	\subfloat[A bench near a grassy area near \textcolor{red}{a parked car}.]{
		\includegraphics[width=0.95\linewidth]{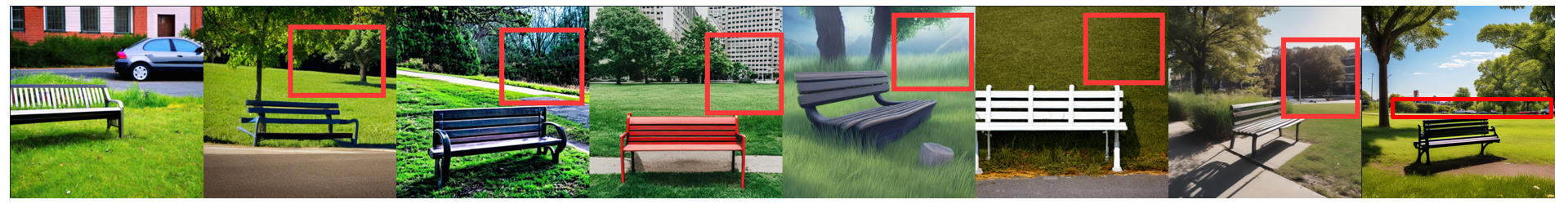}}\\
    \vspace{-2mm}
	\subfloat[\textcolor{red}{A} dog watching \textcolor{red}{a little boy on TV}.]{
		\includegraphics[width=0.95\linewidth]{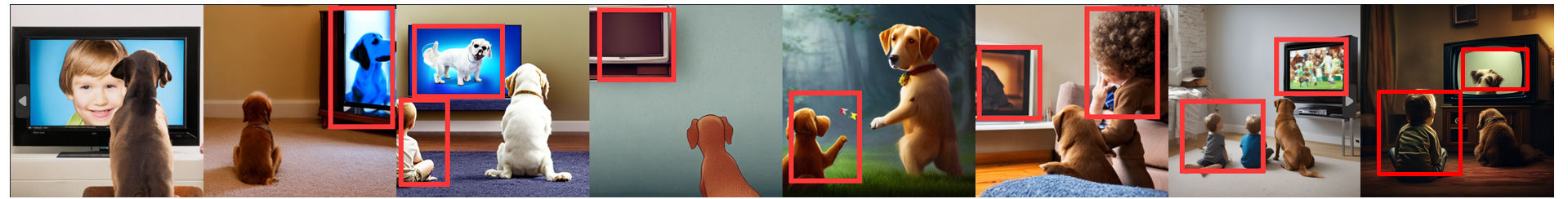}}\\
    \vspace{-2mm}
	\subfloat[A photo of a \textcolor{red}{stop} sign in front of a graffitied \textcolor{red}{truck}]{
		\includegraphics[width=0.95\linewidth]{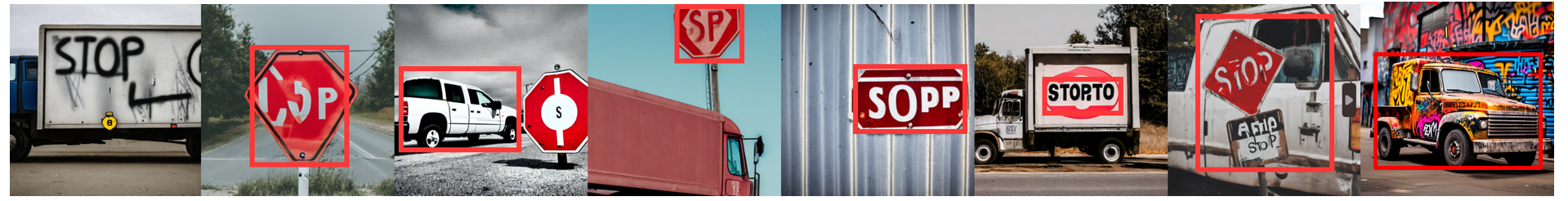}}
	
	\caption{From left to right, respectively: RealignDiff(ours), SD-v1.5~\cite{Rombach_2022_CVPR}, DenseDiffusion~\cite{kim2023dense}, Imagereward~\cite{xu2023imagereward}, Promptist~\cite{hao2022optimizing},  StructureDiffusion~\cite{feng2023trainingfree}, SD-XL~\cite{Rombach_2022_CVPR} and PixArt-$\alpha$~\cite{chen2023pixartalpha}}
	\label{fig:visualize-examples}
\end{figure*}


As shown in Table \ref{tab:refiner}, RealignDiff demonstrates superior performance over the other state-of-the-art (SOTA) methods across all evaluated metrics. It is remarkable that our method's performance slightly surpasses SDXL, even though we used SD v1.5 as the foundational generative model. Specifically, it achieved an FID score of 6.9617 on the MS-COCO dataset, which is significantly lower than those of competing methods, indicating its effectiveness. Additionally, in terms of CLIP and TIFA scores, our method reached 0.3767 and 0.89 on the MS-COCO dataset, respectively. These scores further underscore the ability of RealignDiff to generate semantically coherent and visually compelling images from textual descriptions.

Figures \ref{fig:motivation}, \ref{fig:result} and \ref{fig:visualize-examples} illustrate the qualitative comparison among different methods. It is evident that our RealignDiff achieves the best results in terms of both image quality and semantic consistency between the generated images and the text prompts. ImageReward, Promptist, and StructureDiffusion fail to depict all main objects; SD-v1.5, Midjourney, and DenseDiffusion exhibit misaligned attributes of the objects, such as color.

\begin{figure}[ht]
    \centering
        \includegraphics[height=2.5cm]{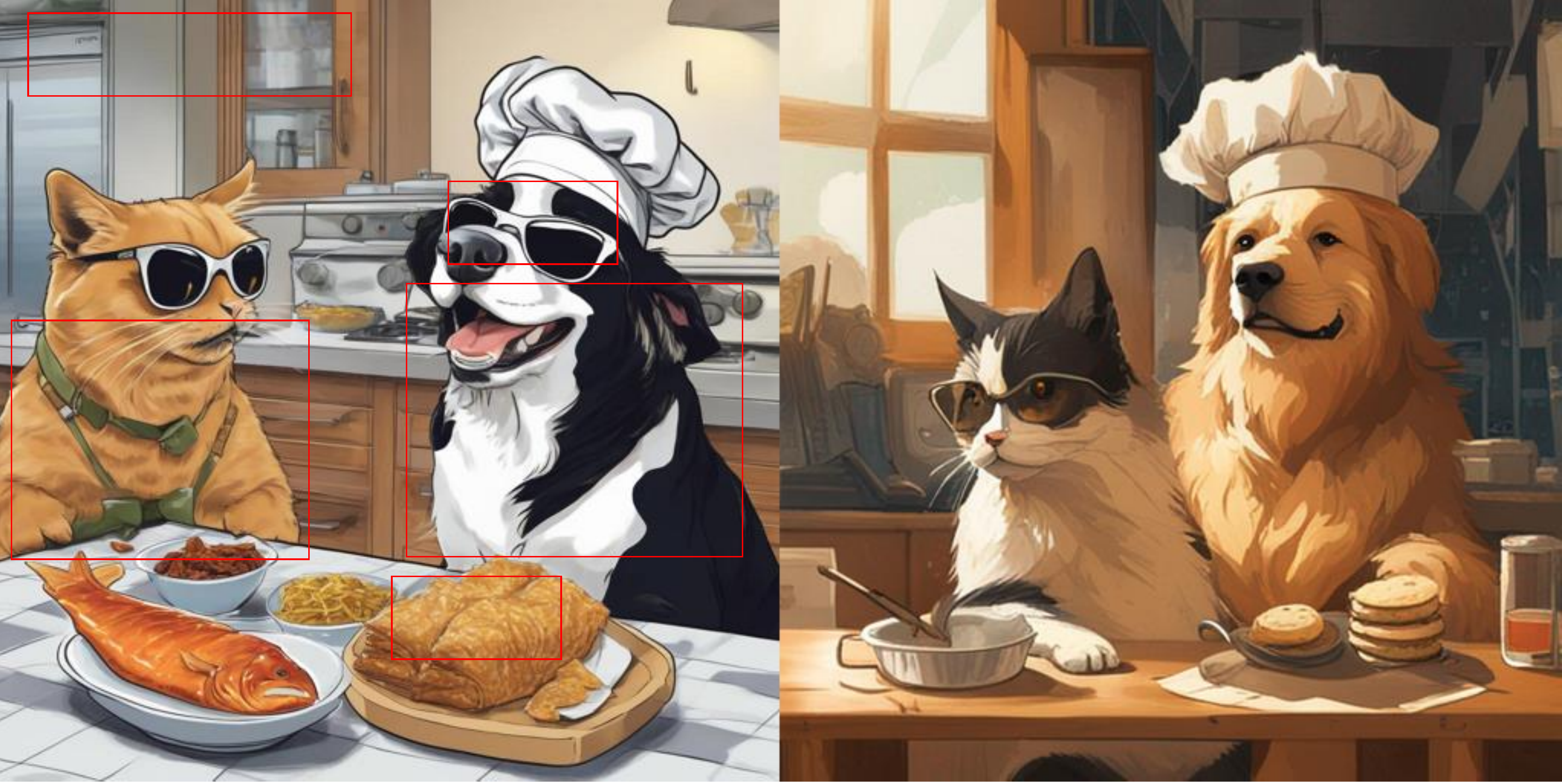} 
        \includegraphics[height=2.5cm]{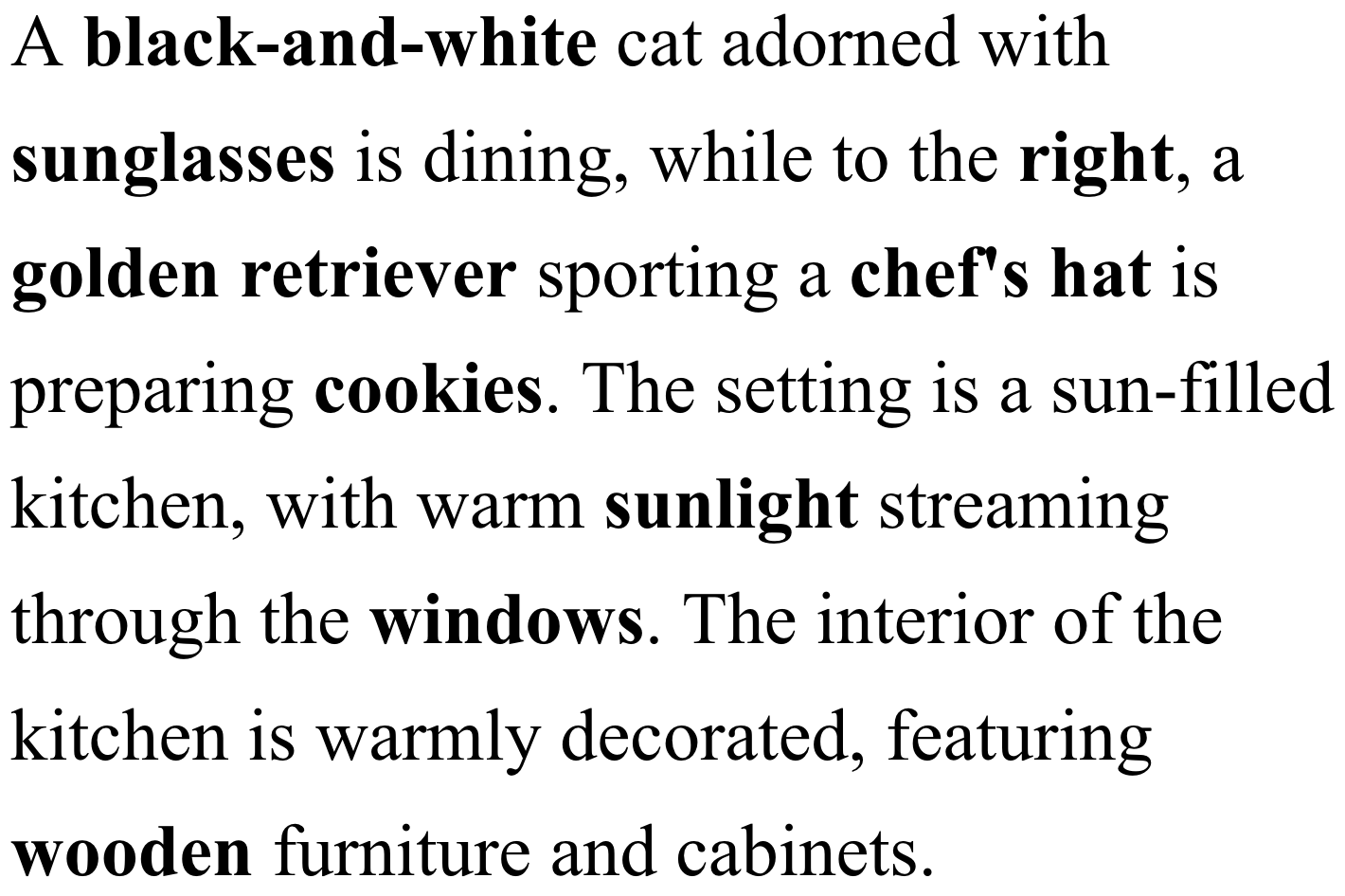} 
    \caption{Generated Image for the complex prompt. Left: SDXL, Right: RealignDiff (Ours)}
    \label{fig:complex_prompt}
\end{figure}

Figure \ref{fig:complex_prompt} presents the generation comparison results for complex prompts.
Compared to SDXL, our RealignDiff can accurately capture the attributes of objects, such as "black and white cat", "golden retriever", "sunlight", "window", and "wooden". Additionally, our method can correctly handle the relative positions of multiple objects, such as "to the right".

\subsection{Ablation Study}
In this section, we first study the effectiveness of the proposed coarse semantic re-alignment and fine semantic re-alignment modules. We then discuss the advantages of our proposed caption reward. Following this, we present a comparison of different LLMs. Finally, we showcase some intermediate generative results and attention maps to illustrate the performance of our approach.

\begin{figure}[ht]
\centering
\includegraphics[width=\linewidth]{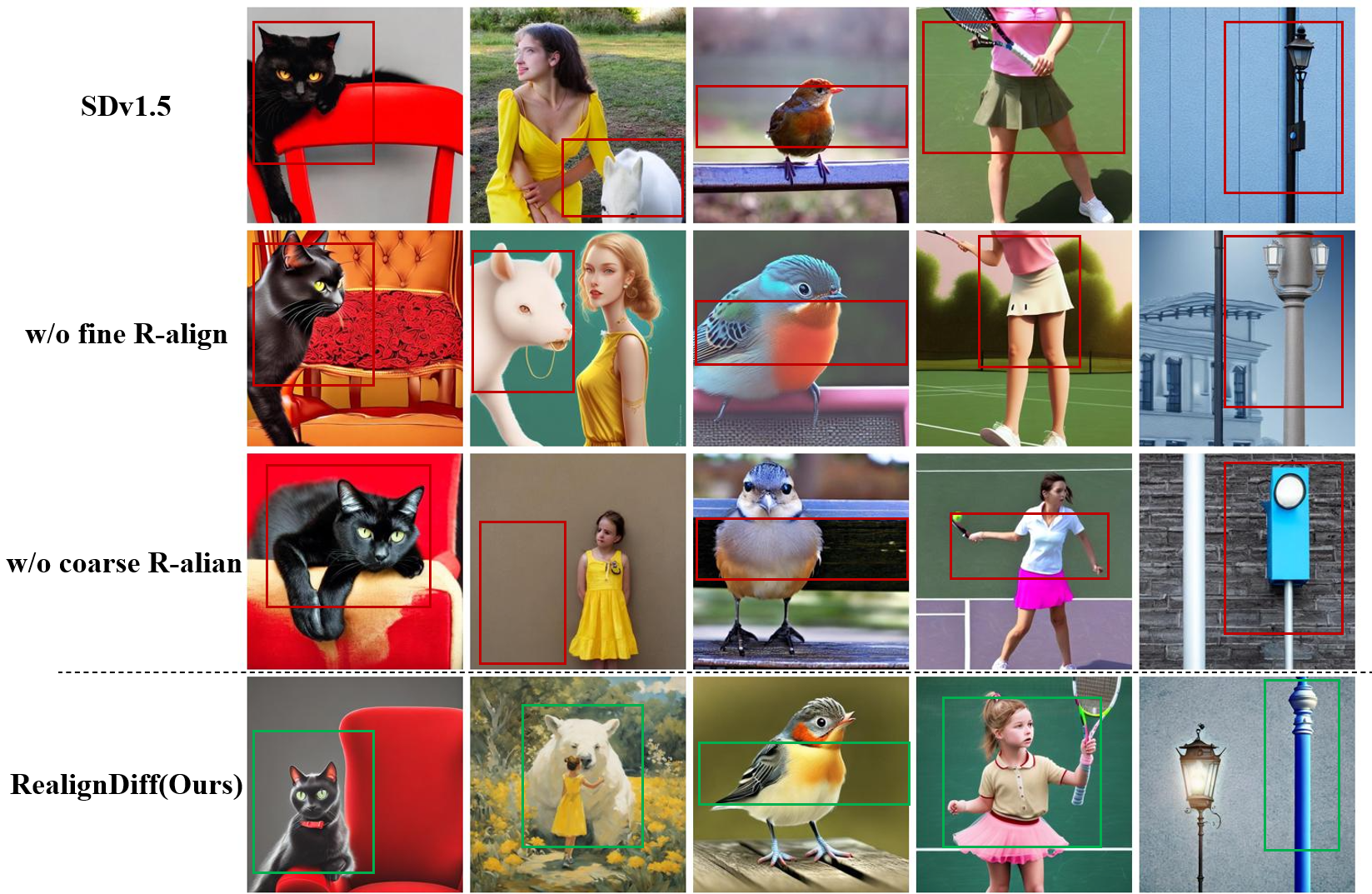}
\caption{Effectiveness of the coarse and fine semantic re-alignment modules. a) a black cat laying on top of the arm of a red chair. b) a girl in a yellow dress and a big white animal. c) a yellow bird is sitting on a park bench. d) a little girl in a khaki shirt and pink skirt playing tennis. e) a street light and a blue pole against a white background}
\label{fig:coarse_to_fine}
\end{figure}

\noindent
\textbf{Coarse \& Fine Semantic Re-alignment.} 
Table \ref{ab1} presents the results of the ablation study for our coarse and fine semantic re-alignment modules on the MS-COCO dataset. The results demonstrate that, in terms of image quality, the coarse semantic re-alignment module reduces the FID from 13.7599 to 7.5349, and the fine semantic re-alignment module further reduces it to 6.9617. With respect to the semantic consistency between the generated images and the input texts, the coarse semantic re-alignment module improves the CLIP score from 0.1626 to 0.2548, and the fine semantic re-alignment module further improves it to 0.3767. These findings suggest that both the coarse and fine semantic re-alignment modules significantly enhance the performance of text-to-image diffusion models.

\begin{table}[ht]
\footnotesize
\centering
\caption{Ablation study of our coarse and fine semantic re-alignment modules on the MS-COCO dataset.}
        \resizebox{0.35\textwidth}{!}{%
\begin{tabular}{cc|ccc}
\hline
Coarse & Fine & FID$\downarrow$ & CLIP$\uparrow$ &TIFA $\uparrow$\\ \hline
& & 13.7599 & 0.1626& 0.75 \\ 
\checkmark & & 7.5349 & 0.2548 & 0.85 \\ 
& \checkmark & 9.3622 & 0.1371 & 0.78\\ 
\checkmark & \checkmark & \textbf{6.9617} & \textbf{0.3767} & \textbf{0.89}\\ \hline
\end{tabular}}
\label{ab1}
\end{table}

Figure \ref{fig:coarse_to_fine} further underscores the effectiveness of the coarse and fine semantic re-alignment modules. The figure reveals that without the coarse semantic re-alignment, the text-to-image diffusion model often fails to capture the main objects mentioned in the text prompts. Without the fine semantic re-alignment, the model struggles to accurately represent the attributes and relationships of the objects described in the input text. However, when both the coarse and fine semantic re-alignment modules are applied, the text-to-image diffusion model is capable of generating high-quality images that are semantically aligned with the input texts.

\noindent \textbf{Caption Reward.}
The reward function is a pivotal component in the coarse semantic re-alignment stage. In this subsection, we evaluate our proposed CaptionReward against other reward functions such as CLIP reward, BLIP reward, and ImageReward. Table \ref{tab:reward_comparison} provides the comparative results among these reward functions on the MS-COCO dataset

\begin{table}[!ht]
\footnotesize
\centering
\caption{Effectiveness of caption reward.}
\resizebox{0.41\textwidth}{!}{%
\begin{tabular}{c|ccc}
\hline
Reward Function & FID$\downarrow$ & CLIP$\uparrow$ &TIFA $\uparrow$\\ \hline
Clip Reward~\cite{hessel2022clipscore} & 14.3091 & 0.1401 & 0.77\\ 
Blip Reward~\cite{li2022blip} & 13.2098 & 0.1400& 0.76\\ 
Image Reward~\cite{xu2023imagereward} & 12.7248 & 0.1587&0.78 \\ 
Caption Reward (Ours) & \textbf{6.9617} & \textbf{0.3767} &\textbf{0.89}\\ \hline
\end{tabular}}
\label{tab:reward_comparison}
\end{table}


\begin{figure}[ht]
    \centering
    \includegraphics[width=\linewidth]{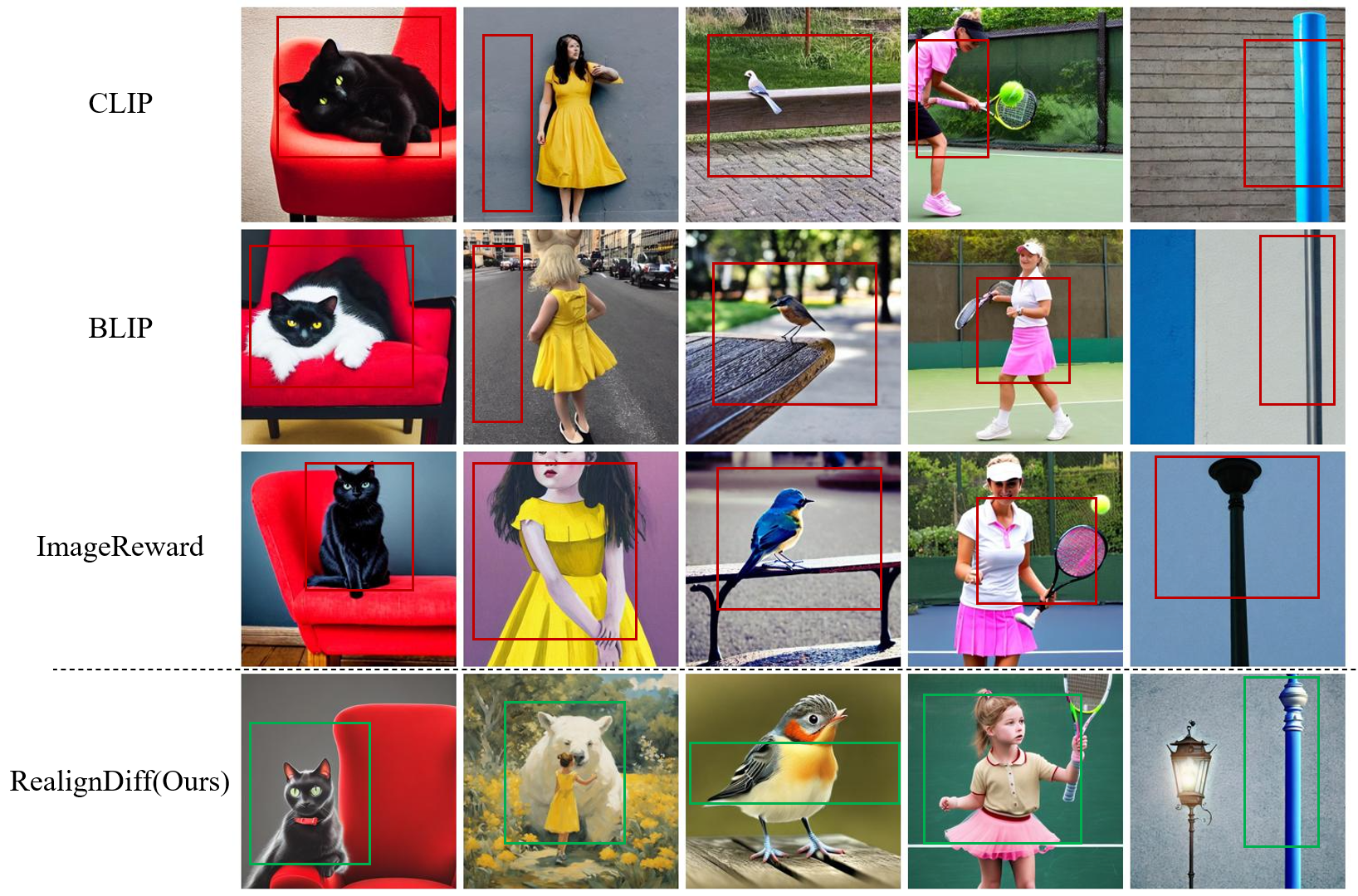}
    \caption{Qualitative Comparison among Reward functions. a) a black cat laying on top of the arm of a red chair. b) a girl in a yellow dress and a big white animal. c) a yellow bird is sitting on a park bench. d) a little girl in a khaki shirt and pink skirt playing tennis. e) a street light and a blue pole against a white background}
    \label{fig:caption_reward}
\end{figure}

As shown in Table \ref{tab:reward_comparison}, our novel Caption Reward outperforms all other reward functions in all metrics, which include CLIP Reward, BLIP Reward, and Image Reward. This superiority can be attributed to CaptionReward's methodology of calculating the reward score by measuring the similarity between the generated caption and the input prompt, rather than measuring the similarity between the generated image and the input prompt, which is the approach taken by the other rewards. The detailed caption provides more nuanced guidance on the appropriateness of the surrounding concepts and context within the image with respect to the given text prompt. Additionally, Figure \ref{fig:caption_reward} further demonstrates the effectiveness of the proposed Caption Reward. As depicted in Figure \ref{fig:caption_reward}, the text-to-image diffusion model, when re-aligned using the Caption Reward, is capable of generating images that are not only of higher quality but also more semantically aligned with the input text than those produced using other reward functions.

\noindent \textbf{Comparison of different LLMs.}
The efficacy of the fine semantic re-alignment module is intrinsically linked to its ability to tag images accurately and modulate attributes effectively through large language models (LLMs). This section aims to delve into an ablation study that analyzes the success rates of various image tagging models and LLMs on the ViLG-300 dataset, including ChatGPT, GPT-4~\cite{Achiam2023GPT4TR}, Vicuna-7b~\cite{vicuna2023}, and Llama2-7b~\cite{Touvron2023Llama2O}. 

\begin{table}[ht]
\footnotesize
\centering
\caption{The success rates of different image tagging and large language models in the local dense caption generation module on the ViLG-300 dataset.}
\resizebox{0.47\textwidth}{!}{%
\begin{tabular}{c|cccc}
\hline
Model  & Llama2-7b & Vicuna-7b  & ChatGPT & GPT-4\\ \hline
RAM~\cite{zhang2023recognize}   & 63\% & 81\%  & 92\% & \textbf{99\%} \\
Tag2Text~\cite{huang2023tag2text} &51\% & 76\% &85\% &91\%  \\
\hline
\end{tabular}}
\label{tab:success_rates}
\end{table}

Table \ref{tab:success_rates} showcases the success rates of both RAM and Tag2Text when paired with the aforementioned LLMs. It is noteworthy that the success rate of RAM+GPT-4 can achieve 99\%. The findings indicate: 1) ChatGPT, while slightly lagging behind GPT-4, presents promising outcomes, especially with RAM. This underscores the versatility and robustness of the ChatGPT model; 2) On the other end of the spectrum, Llama2-7b exhibits the lowest success rates with both tagging models. This could hint at possible areas of refinement or potential incompatibilities between the tag model and LLM. In the future, we can improve performance and save costs by specifying fine-tuning of the task.

\begin{figure}[ht]
    \centering
     \includegraphics[width=\linewidth]{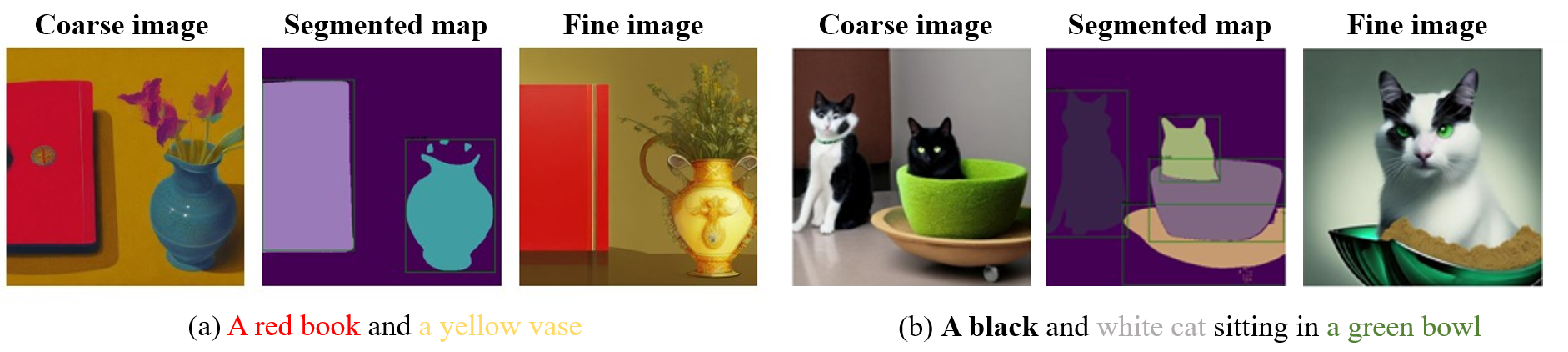}
    \caption{Intermediate generative results of RealignDiff.}
    \label{fig:CTF}
\end{figure}

\noindent \textbf{Intermediate generative results and attention maps.}
Figure \ref{fig:CTF} presents the intermediate generative results of RealignDiff, displaying the progression from left to right: the initial coarse images following coarse semantic re-alignment, the segmented maps derived from the coarse images, and the final high-quality, semantically-aligned images produced after fine semantic re-alignment, which achieve fine-grained attribute binding.
In Figure \ref{fig:CTF}(a), the \textbf{vase} undergoes a reassignment of its attribute to \textbf{yellow}. In Figure \ref{fig:CTF}(b), the \textbf{quantity of cats} is reassigned to \textbf{one}, their \textbf{location} is redefined as \textbf{on the bowl}, and their \textbf{color} is reassigned to a \textbf{black-and-white} pattern.

\begin{figure}[ht]
	\centering  
	\subfloat[]{
		\includegraphics[width=0.24\linewidth]{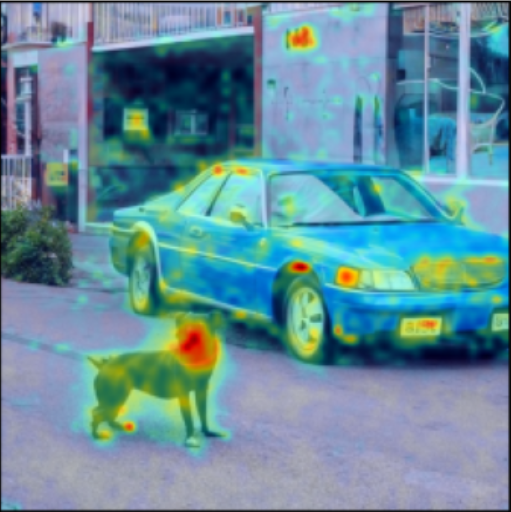}} 
	\subfloat[]{
		\includegraphics[width=0.24\linewidth]{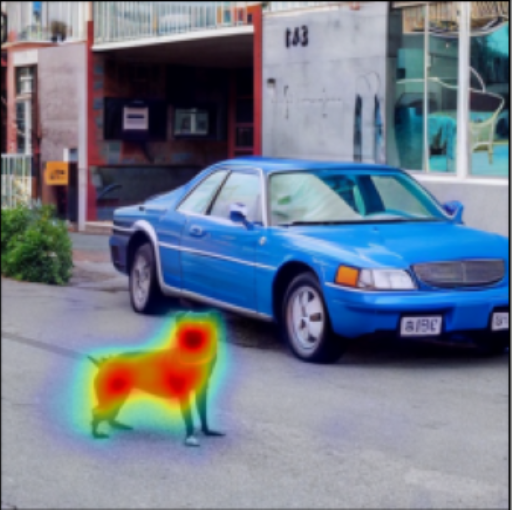}} 
  \subfloat[]{
		\includegraphics[width=0.24\linewidth]{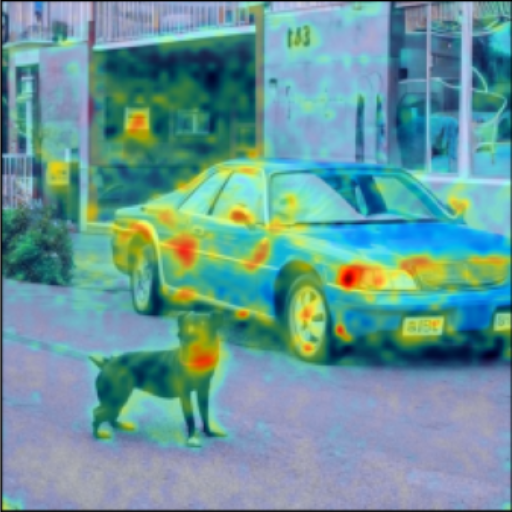}} 
	\subfloat[]{
		\includegraphics[width=0.24\linewidth]{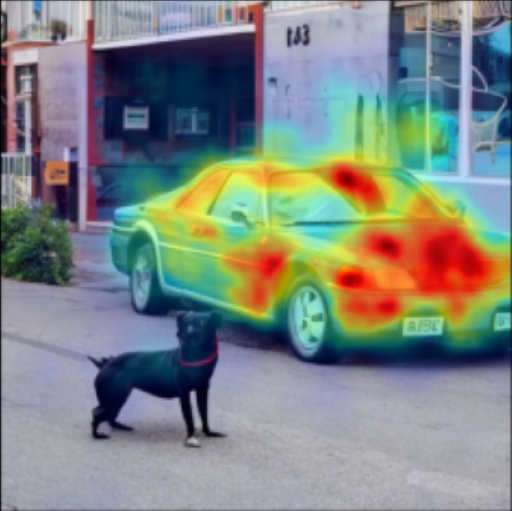}} 
	\caption{Intermediate attention maps of "A black dog on the street next to a blue car." (a) 'black' before re-align, (b) 'black' after re-align, (c) 'blue' before re-align, (d) 'blue' after re-align}
	\label{fig:map}
\end{figure}

We use Diffusion Attentive Attribution Maps (DAAM)~\cite{tang2023daam} to visualize the intermediate attention maps of key attribute tokens before and after fine semantic re-alignment.  As displayed in Figure~\ref{fig:map}, the color attributes \textbf{black} and \textbf{blue} get more attention in the designated regions after fine semantic re-alignment, leading to better alignment and generative performance.

\section{Conclusion}
\label{sec:conclusion}

In this paper, we propose a novel two-stage coarse-to-fine semantic re-alignment method, RealignDiff, to enhance the alignment between descriptions and corresponding images within the text-to-image diffusion models.
The initial coarse semantic re-alignment stage entails fine-tuning the text-to-image model from a global semantic perspective. This stage is crucial for ensuring that the generated images faithfully depict the objects and entities described within the given textual input. 
The fine semantic re-alignment stage occurs without the need for additional training data, allowing for the accurate capture of object attributes and relationships. 
Experimental results on MS-COCO and ViLG-300 datasets demonstrate that RealignDiff outperforms other baselines in terms of both visual quality and semantic similarity with input prompt. 

\noindent
\textbf{Limitations and future works}
In the fine semantic re-alignment stage, if the large language model fails to provide accurate intermediate results, it may hinder the refinement of previously generated images. Future work will focus on overcoming this limitation. Additionally, we aim to explore dynamic learning from multiple reward functions, such as semantic and aesthetic, within the diffusion model.

\ifCLASSOPTIONcaptionsoff
  \newpage
\fi

\begin{bibliographystyle}{IEEEtran}
\begin{bibliography}{main}
\end{bibliography}
\end{bibliographystyle}

\end{document}